\documentclass[letterpaper, 10 pt, conference]{ieeeconf}  

\IEEEoverridecommandlockouts                              

\usepackage{epsfig}
\usepackage{times}
\usepackage{amsmath}
\usepackage{amssymb}
\usepackage{graphicx}
\usepackage{cite}
\usepackage{subcaption}
\usepackage[]{algorithm2e}
\usepackage{framed}

\usepackage{multirow} 
\usepackage{booktabs} 

\title{\LARGE \bf
An Insect-scale Untethered Laser-powered Jumping Microrobot
}

\author{Palak Bhushan$^{*}$ and Claire Tomlin$^{*}$
\thanks{$^{*}$The authors are with the Department of EECS, University of California, Berkeley, CA 94720, USA.
        {\tt\small palak@berkeley.edu} \text{(corresponding author)}, {\tt\small tomlin@berkeley.edu}.}%
}

\begin{document}

\maketitle
\thispagestyle{empty}
\pagestyle{empty}

\begin{abstract} 
We present the design of an insect-sized jumping microrobot measuring 17mm$\times$6mm$\times$14mm and weighing 75 milligrams. The microrobot consumes 6.4mW of power to jump up by 8mm in height. The tethered version of the robot can jump 6 times per minute each time landing perfectly on its feet. The untethered version of the robot is powered using onboard photovoltaic cells illuminated by an external infrared laser source. 
It is, to the best of our knowledge, the lightest untethered jumping microrobot with onboard power source that has been reported yet. 
\end{abstract} 

\section{Introduction} 

Milligram-scale microrobots (or, $\mu$bots) can utilize a variety of locomotion strategies to navigate around the world, including crawling, rolling, walking, jumping and flying. 
Crawlers are less robust to the ever changing environment compared to rollers and walkers, followed by jumpers \cite{locomotion08}, with flyers being the most robust among these. 
Robustness generally increases the less the bot interacts with the environment especially with its multitude of surfaces. 
Jumpers and flyers minimize this interaction by jumping/flying over the obstacles and to their next destination. 

Locomotion energetics generally follows an inverse trend from above with flight, and especially hovering, being the most demanding \cite{actuator_selection}. 
It is then no surprise that only two flying $\mu$bots have been reported yet that can lift-off without tethers \cite{robofly18, xwing19}. 
Jumping, even though less demanding than flight, is energetically more demanding than walking and poses its own challenges. 

Untethered jumping has been demonstrated before using mechanical and chemical approaches. 
\cite{churaman11} reports an 8mg spring-mass device that can jump up by 32cm by rapidly releasing mechanical energy stored in a spring, but the device needs to be loaded and released manually and has no actuators. 
\cite{churaman12} reports a 300mg bot that can jump up by 8cm using explosive chemical energy, but the capacitors in the igniting circuit needs to be charged manually for each jump. 
Another 34mg jumper reported in \cite{hotplate} jumps up by 30cm but it requires a controlled environment, namely a hot plate, to transfer energy to its SMA actuators in order to jump. 
Apart from the needed manual intervention and/or a controlled environment, the downside is that all these bots can jump only once. 
Note that even if the chemical jumper had multiple independent chemical packets to release the explosive energy multiple times, the jumps won't be repeatable indefinitely with onboard fuel sources being consumed up after some time. 

This work is inspired from the silicon jumping $\mu$bots reported in \cite{greenspun18} which being electrically powered can in principle jump indefinitely as long as they have power (say, using solar cells). These weight around 43mg, are monolithic, and can jump up by a millimeter while being driven by an external power supply. The electrostatic actuators used though demand 100V to operate. 
The lack of good voltage step-up $\mu$-circuits, along with the need to use complex control signals to control their multiple actuators in sync, has prevented their tetherless operation as of now. 

We go the electrically-powered route as well and use electromagnetic (EM) actuation due to their ease in fabrication and low-voltage operation. EM actuators have been used before in $\mu$bots \cite{rolling_ral19, baybug18, baybug19}, but not for the jumping ones. 
We choose our coil impedance in order to power it directly using 1mg photovoltaic (PV) cells eliminating the need for any voltage conversion circuitry. 
In order to simplify fabrication and control, we make our design operate using a single actuator by making other functions occur passively. 

\begin{figure}
\centering
\epsfig{file=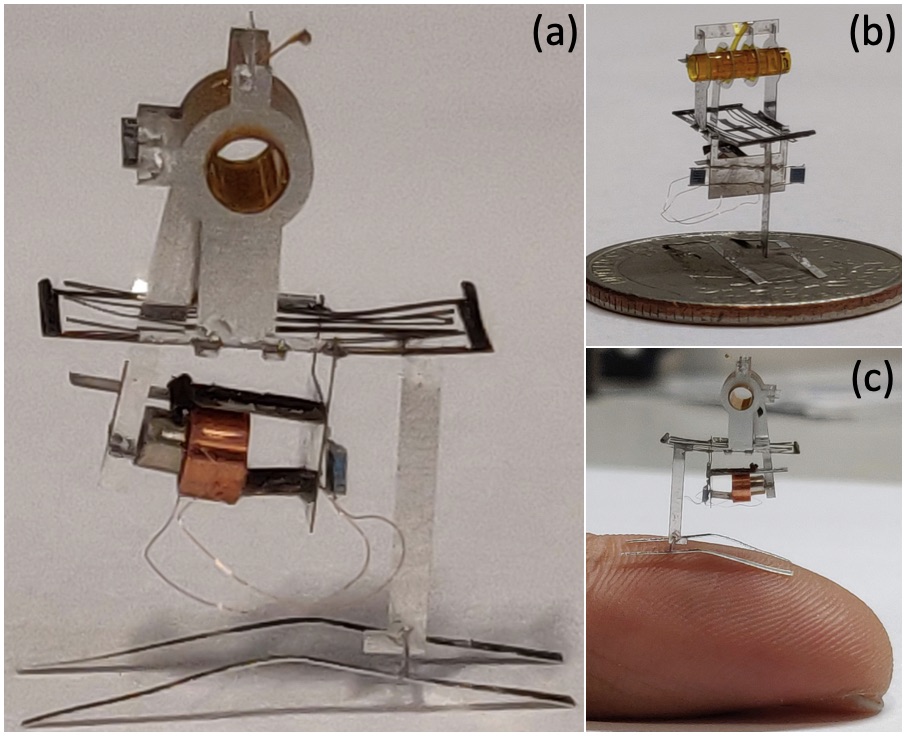,width=3.4in}
\vspace{-1.9em}
\caption{\small{(a) Jumping $\mu$bot, compared with (b) a quarter dollar, and, (c) an index finger.}}
\vspace{-2.1em}
\label{fig:main-pic}
\end{figure}

\section{Design} 

\subsection{Underlying Principle} 

\begin{figure}[h]
\vspace{-1em}
\centering
\epsfig{file=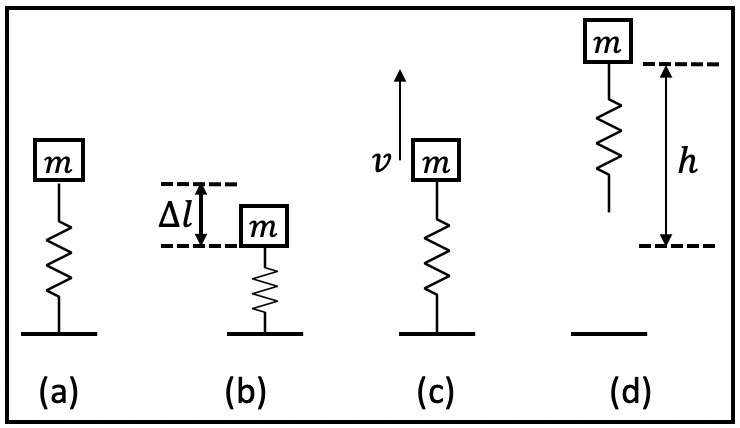,width=2.6in}
\vspace{-0.6em}
\caption{\small{Underlying principle of the jumping bot. (a) Neutral state of the spring. (b) Spring deflected by the maximum amount storing potential energy. (c) Released spring just before loosing ground contact. (d) Bot at the maximum jump height.}}
\vspace{-1.8em}
\label{fig:principle}
\end{figure}

The principle behind is a spring that stores potential energy by getting pulled by the onboard motors (see Fig. \ref{fig:principle}). Rapidly releasing the spring releases the stored energy which then causes the $\mu$bot to jump up to a certain height. 
Neglecting air resistance and using energy conservation we have  
$\frac{1}{2} k \Delta l^2 = mgh $, 
where $m$ is the net mass of the $\mu$bot, $\Delta l$ is the maximum spring deflection, and $h$ is the jump height measured from the maximum deflected state. As a reference, a 1cm jump of a 100mg $\mu$bot should require $\approx$ 10$\mu$J of spring energy. 

\subsection{Spring design} 

\begin{figure}[h]
\vspace{-1.0em}
\centering
\epsfig{file=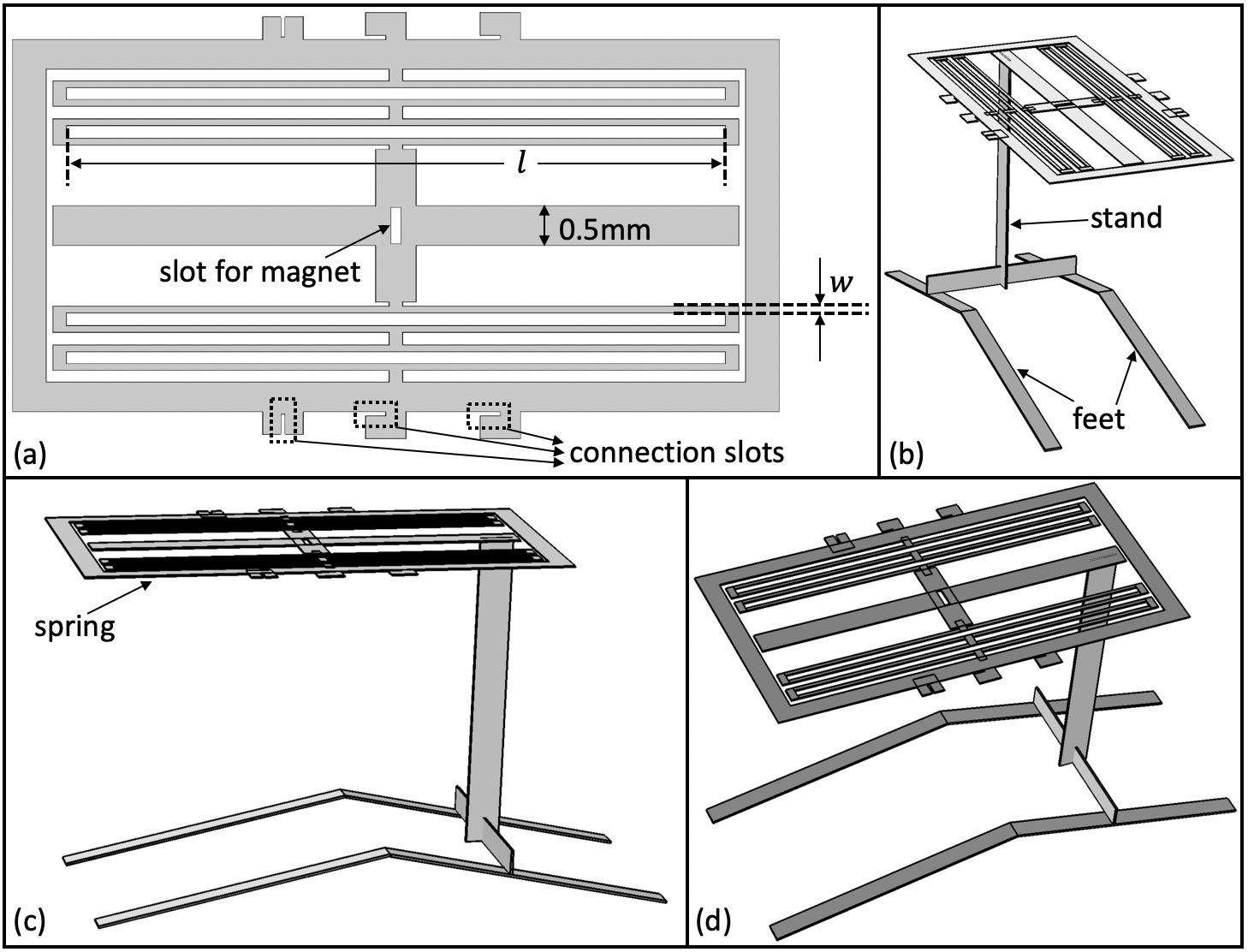,width=3.4in}
\vspace{-1.7em}
\caption{\small{(a) Planar spring design showing connection slots later used to attach parts perpendicularly to the spring. (b), (c) \& (d) Different views of the stand and feet glued to the spring. }}
\vspace{-1.6em}
\label{fig:spring}
\end{figure}

\begin{figure}[h]
\centering
\epsfig{file=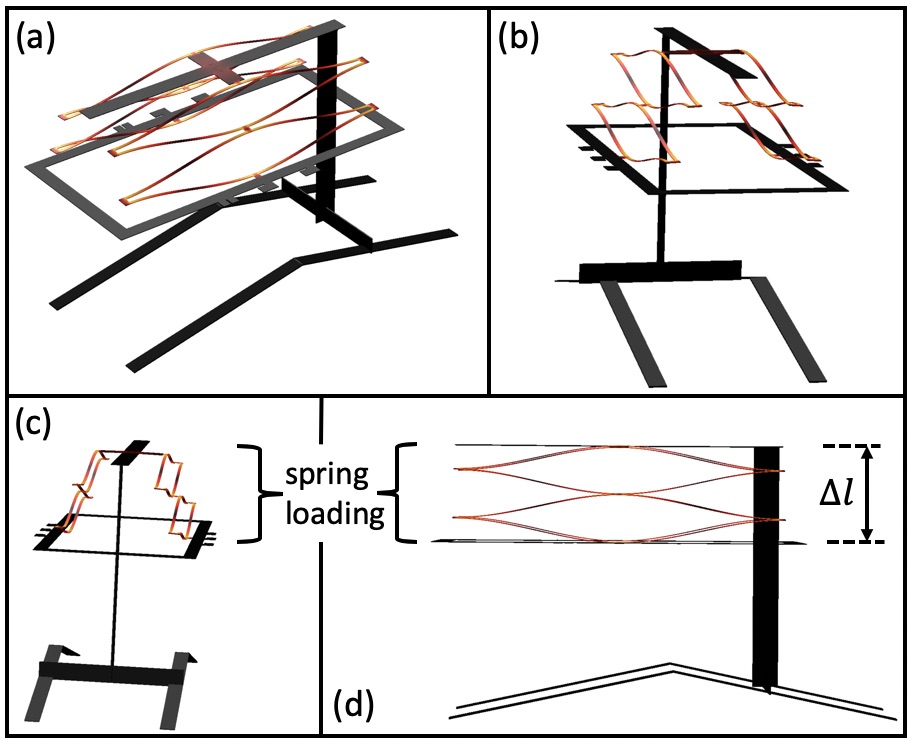,width=3.2in}
\vspace{-0.7em}
\caption{\small{Different views of the spring at full deflection (FEA). }}
\vspace{-0em}
\label{fig:spring-def}
\end{figure}

The planar spring design shown in Fig. \ref{fig:spring}(a) is laser cut from a 25.4$\mu$m-thick Stainless Steel sheet. The individual beam length $l$ and width $w$ are tuned using finite element analysis (FEA) to achieve a desired stiffness $k$ while allowing for the maximum out-of-plane deflection $\Delta l$ (that is, keeping maximum strains below the Yield strain). 
A T-shaped stand is glued perpendicularly to the spring plane, and then two feet are glued on to the stand (see Fig. \ref{fig:spring}(b)). Both the stand and the feet are laser cut from a 50$\mu$m-thick Aluminum sheet. 
The spring can be seen in its fully deflected state in Fig. \ref{fig:spring-def} which is computed via FEA simulations. 

\subsection{Spring loading mechanism} 

\begin{figure}[h]
\vspace{-0.8em}
\centering
\epsfig{file=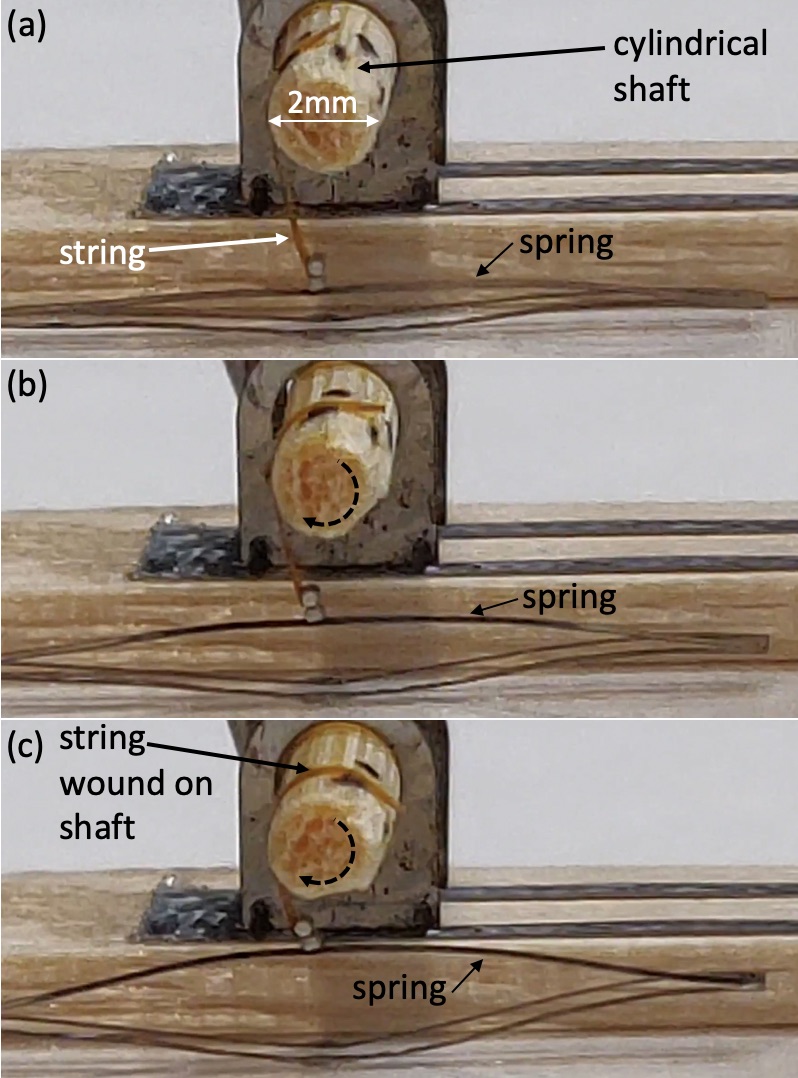,width=2.8in}
\vspace{-0.5em}
\caption{\small{Spring loading mechanism tested on a simple leaf spring. }}
\vspace{-2em}
\label{fig:spring-loading}
\end{figure}

The spring is pulled up, or loaded, via a 12.7$\mu$m-thick 0.5mm-wide Kapton string attached to the center of the spring. 
This pull is obtained by winding the string around a clockwise rotating cylindrical shaft as seen in Fig. \ref{fig:spring-loading}.

\subsection{Spring release mechanism} 

\begin{figure}[h]
\vspace{-0.9em}
\centering
\epsfig{file=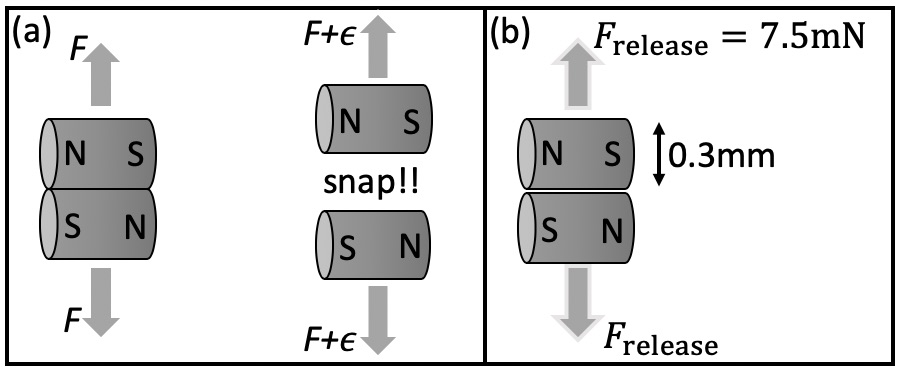,width=3.0in}
\vspace{-0.6em}
\caption{\small{Passive spring release concept using magnets. }}
\vspace{-0.5em}
\label{fig:spring-release}
\end{figure}

We want our jumping $\mu$bot to operate using a single motor. That motor will be used to rotate the shaft, so our spring release mechanism should not be controlled by a separate motor but instead happen passively. We use magnets to create our automatic release mechanism (see Fig. \ref{fig:spring-release}(a)). Two anti-parallel magnets placed side by side are attracted by a certain amount of force $F$. If opposing forces acting on the magnet exceed even a little beyond this attractive force, this leads to an instability and the magnets are released rapidly. 
The magnets used are Neodymium grade N52 of diameter $=0.3$mm and height $=0.5$mm, and their release force is measured at $F_{\text{release}}=7.5$mN (see Fig. \ref{fig:spring-release}(b)). 

\begin{figure}[h]
\vspace{-1.4em}
\centering
\epsfig{file=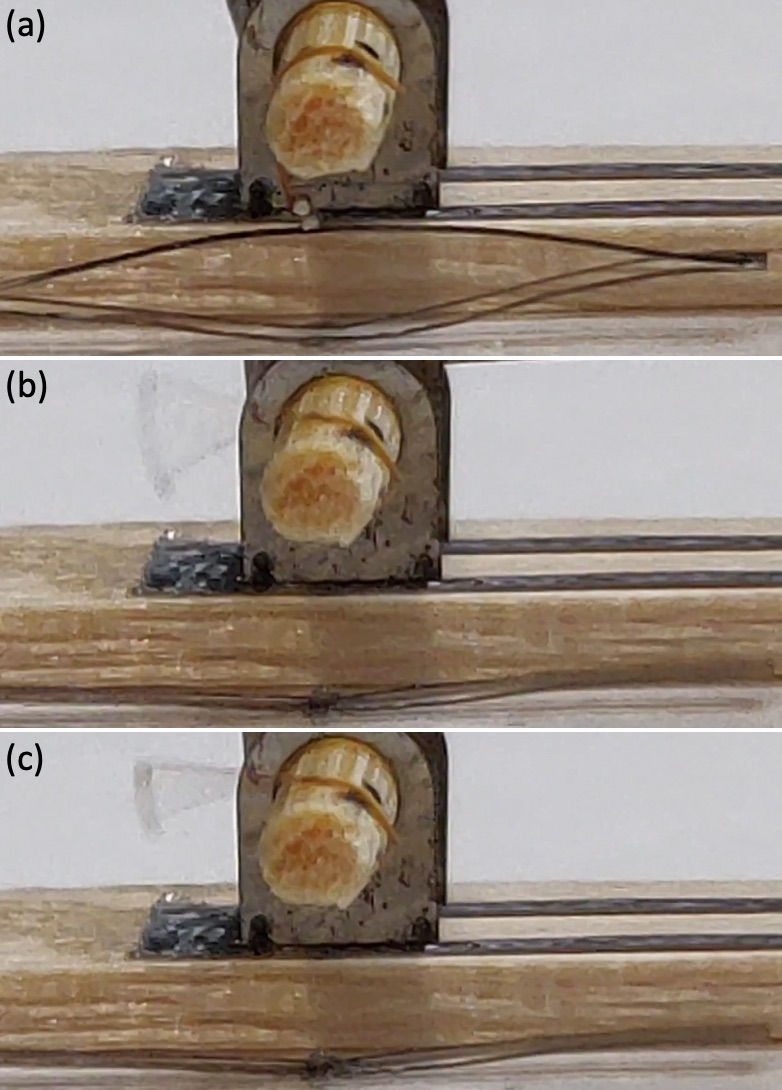,width=2.8in}
\vspace{-0.5em}
\caption{\small{Passive spring release in action. (a) Spring at maximum deflection just before the magnets snap. (b) String vibrating just after the release. (c) String vibrations dampening over time. }}
\vspace{-1em}
\label{fig:spring-release-real}
\end{figure}

Again, the spring is loaded by winding the string but this time the string connects to the spring via these magnets. One magnet is glued to the center of the spring in the slot shown in Fig. \ref{fig:spring}(a), and another is glued to the end of the string. The opposing forces on the magnets increase with increasing spring deflection and at the correct threshold the magnets snap (see Fig. \ref{fig:spring-release-real}). 
The release force informs our spring stiffness and deflection choice so that it is storing around 10$\mu$J at the time of release. Choosing $k=2.5$N/m $\Rightarrow \Delta l = \frac{\text{7.5mN}}{\text{2.5N/m}}=3$mm at the time of release $\Rightarrow$ a stored energy of $\frac{1}{2} \cdot$2.5N/m$\cdot$(3mm)$^2=11.25\mu$J at the time of release.

\subsection{Shaft rotation mechanism} 

In order to wind the Kapton string to generate the pull, the shaft should be made to rotate in a clockwise direction. 
However, it is difficult to build a rotary motor at the milligram scale. Hence the approach we take is to convert the motion of an easy-to-build linear oscillatory actuator to a continuous rotation motion. We do so by using ratchets as is also described in \cite{rolling_ral19} in more detail. 

\subsubsection{Micro-ratchet} 

\begin{figure}[h]
\centering
\epsfig{file=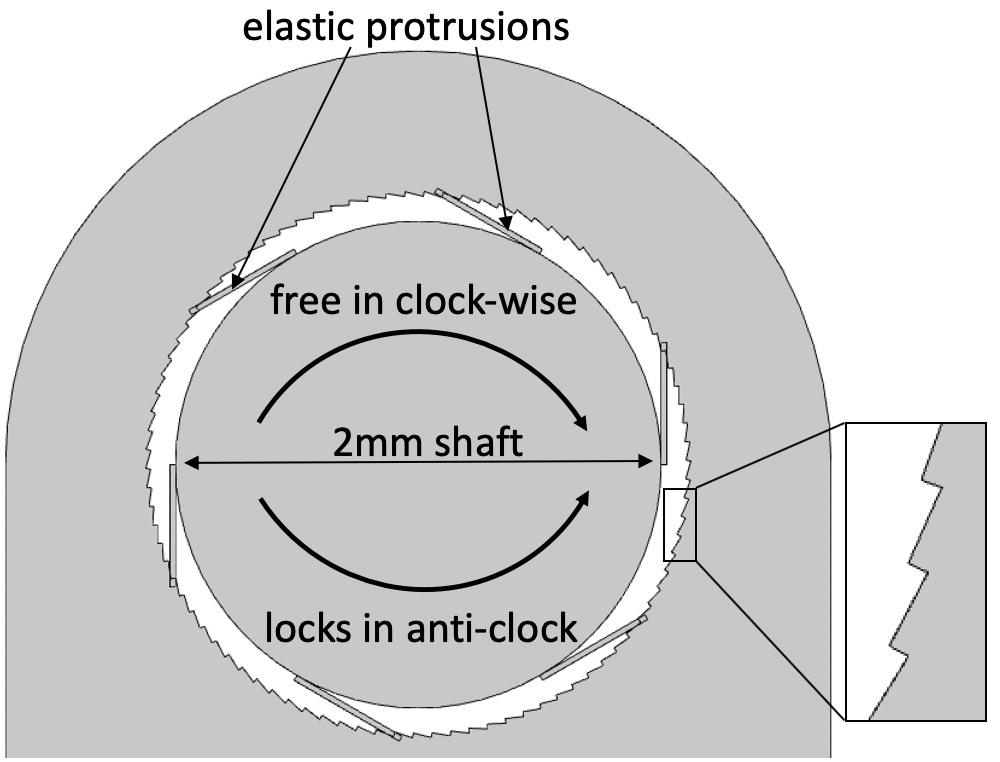,width=2.2in}
\vspace{-0.8em}
\caption{\small{Cross-section of a micro-ratchet mechanism made using flexible beams on a shaft and a patterned hole. The peaks in the pattern are 25$\mu$m high and are spaced 4$^\circ$, or, $\approx$ 70$\mu$m apart. }}
\vspace{-1.8em}
\label{fig:ratchet}
\end{figure} 
The inner shaft of the ratchet (shown in Fig. \ref{fig:ratchet}) is free to rotate relative to the outer ring when it is rotated in a clockwise direction. In this condition, the elastic protrusions emanating from the shaft slide over the zig-zag patterns on the inner perimeter of the ring. These elastic beams are bent by 25$\mu$m more (in addition to any pre-deflection) when encountering the peaks in the pattern. When rotated anti-clockwise, the shaft locks relative to the ring. In this reverse operation, the elastic beams push the falling edge of the pattern head-on, and motion can only be achieved if the beams buckle. This buckling requires orders of magnitude higher torque compared to the simple sliding motion from before and this configuration can be considered as locked for the purposes of this paper. 

\begin{figure}[h]
\vspace{-0.8em}
\centering
\epsfig{file=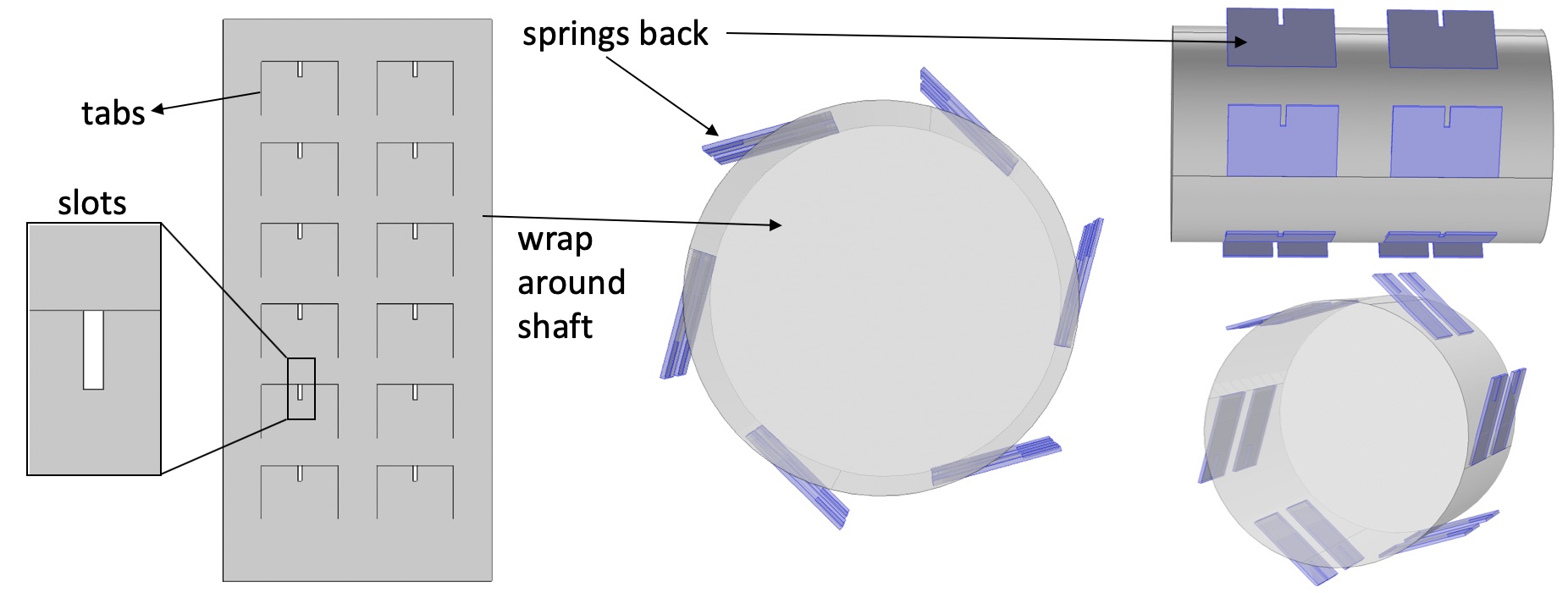,width=3.4in}
\vspace{-1.7em}
\caption{\small{Fabrication of the shaft for the micro-ratchet mechanism. 60$^\circ$ spaced flexible beams are obtained by wrapping a laser cut Kapton sheet with tabs on to a Kapton tube.}}
\vspace{-0.5em}
\label{fig:shaft}
\end{figure}
Fig. \ref{fig:shaft} shows the construction of the inner shaft. 12 tabs are laser-cut on a 12.7$\mu$m-thick Kapton sheet. This laser-cut sheet is then rolled on to a 2mm-diameter Kapton tube and glued in place. The rest of the sheet adheres to the curved surface of the tube due to the glue, but the unglued tabs retain their planar shape thus acting as our desired elastic protrusions. 

\begin{figure}[ht]
\vspace{-1.2em}
\centering
\epsfig{file=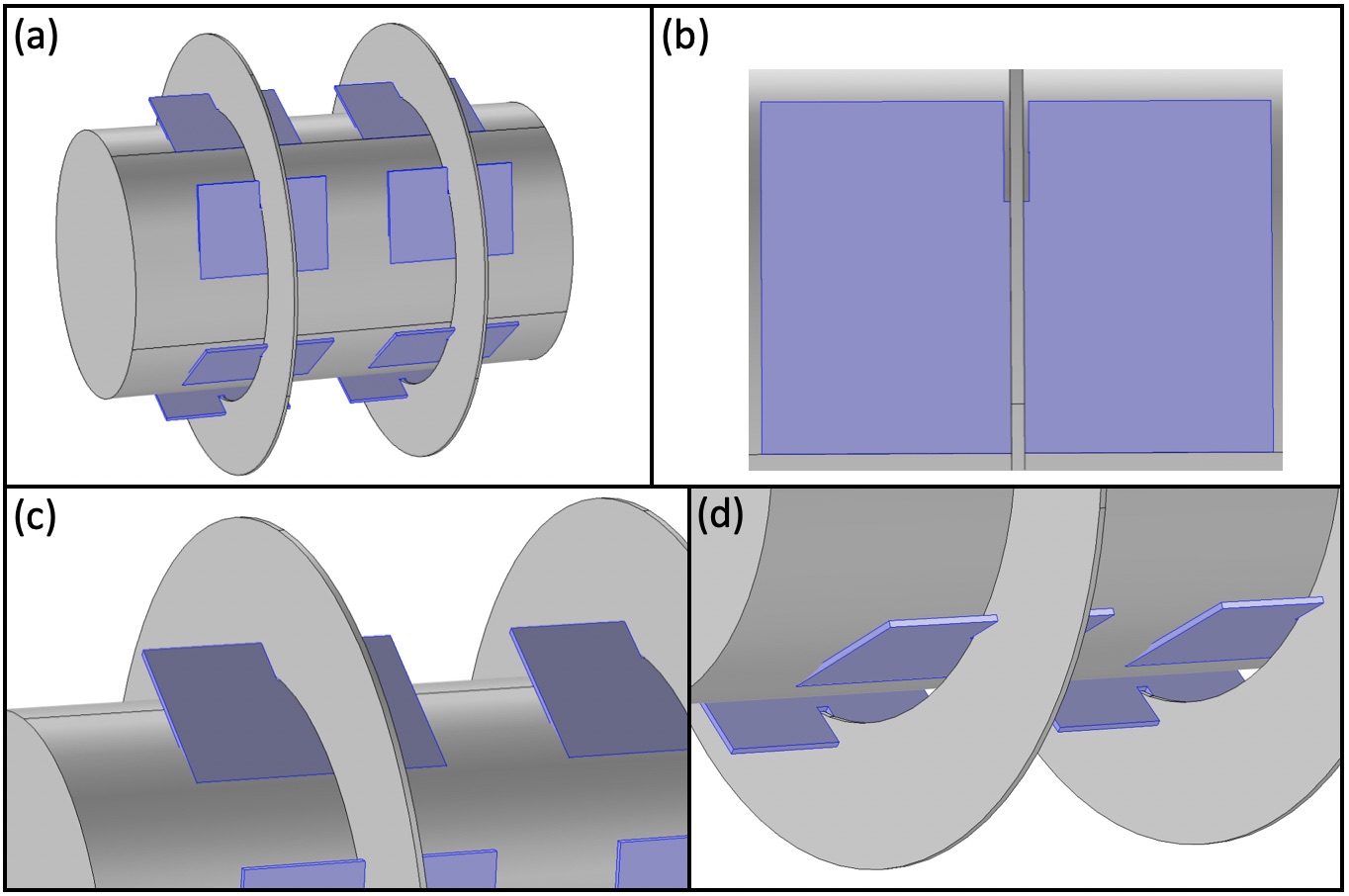,width=3.0in}
\vspace{-0.6em}
\caption{\small{(a) Laser-cut patterned steel rings are slid in to the shaft such that (b) the ring passes through the slots in each of the tabs/elastic beams. (c) \& (d) show a better view of the same. }}
\vspace{-1.0em}
\label{fig:ring-shaft}
\end{figure}
Rings with patterned holes are laser cut using 50$\mu$m-thick Aluminum. These rings slide into the slots previously cut in each of the tabs as seen in Fig. \ref{fig:ring-shaft}. The slots prohibit any sideways motion of the rings thus keeping them in place. 

\subsubsection{Double-ratchet} 

\begin{figure}[h]
\centering
\epsfig{file=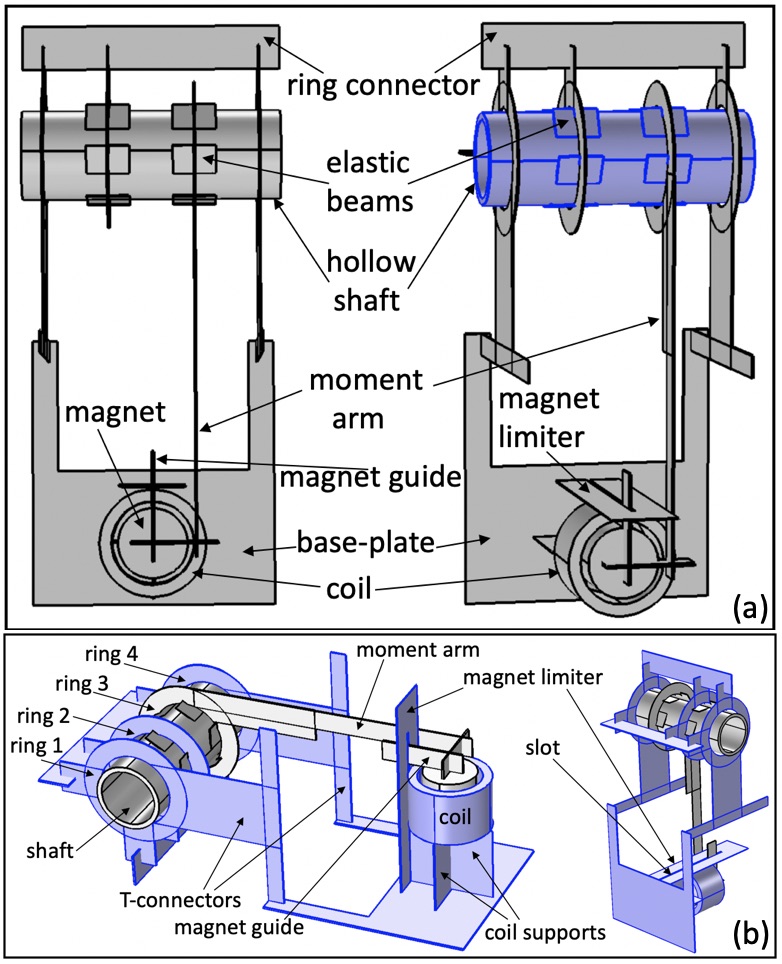,width=3.4in}
\vspace{-1.7em}
\caption{\small{The double ratchet mechanism used to produce continuous rotation motion. (a) Shaft colored in blue. (b) All parts acting as one rigid part colored in blue. }}
\vspace{-2em}
\label{fig:double-ratchet3}
\end{figure}

\begin{figure}[h]
\vspace{-0.7em}
\centering
\epsfig{file=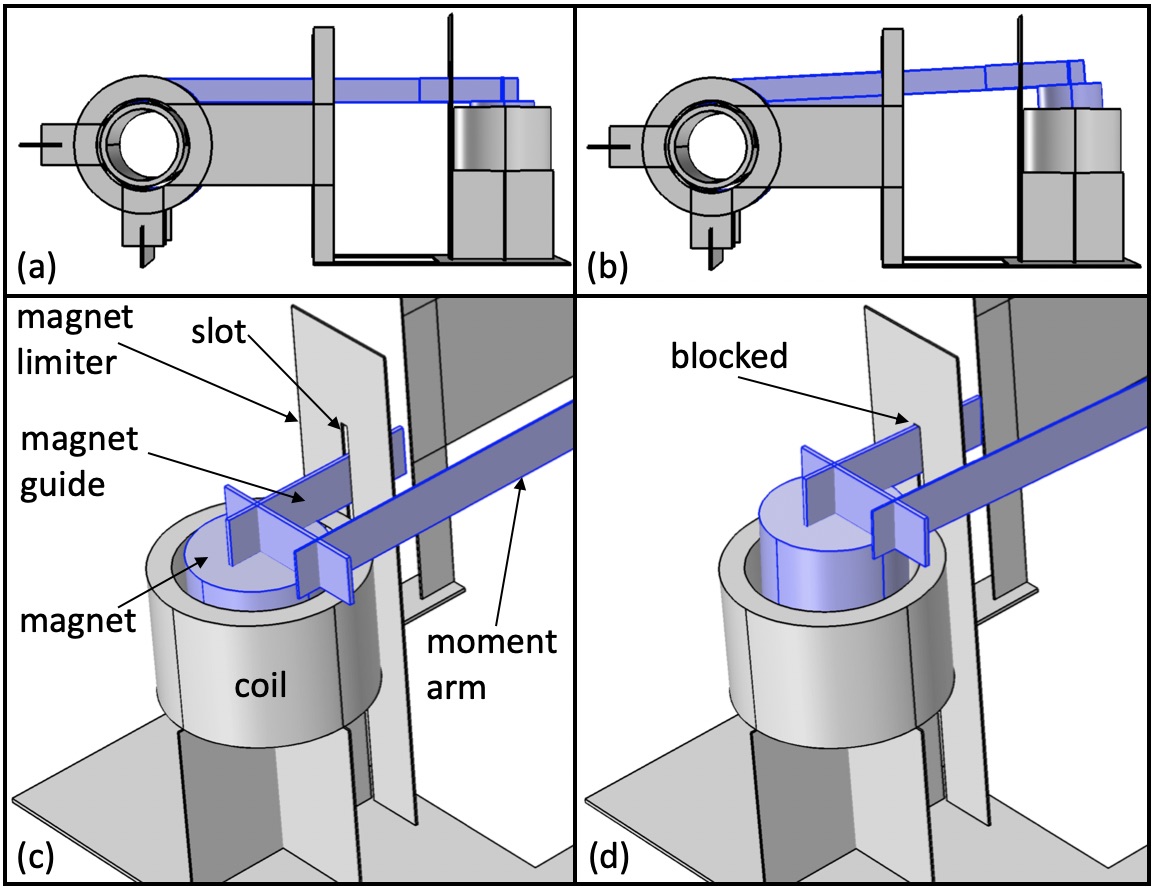,width=3.4in}
\vspace{-1.7em}
\caption{\small{(a) \& (b) Two extreme positions of the moment arm separated by $\approx 2^\circ$ rotation. (c) \& (d) Magnet limiter in action. }}
\vspace{-1em}
\label{fig:magnet-limiter}
\end{figure}

All planar parts in Fig. \ref{fig:double-ratchet3} are laser cut from a 50$\mu$m-thick Aluminum sheet. Four rings are used in this mechanism, with rings 1 \& 4 supporting the shaft, and rings 2 \& 3 acting as ratchets. Rings 1, 2 \& 3 are rigidly connected to each other using two rectangular beams, called ring connectors. The two ring connectors can be seen rotated 90$^\circ$ relative to one another. Two T-shaped connectors connect rings 1 \& 4 to the U-shaped base-plate. Two rectangular beams, called coil supports, are glued perpendicularly to the base-plate and then a coil is glued above them. A rectangular magnet limiter, with a narrow 100$\mu$m slot in it, is also glued perpendicularly to the base-plate and touching the coil. The magnet is concentric to the coil in its neutral position and is glued to a rectangular magnet guide (see Fig. \ref{fig:double-ratchet3} \& Fig. \ref{fig:magnet-limiter}(c)). A long moment arm emanating from ring 3 connects to the magnet guide via a small orthogonal rectangular beam (see Fig. \ref{fig:magnet-limiter}(c)). 

Input motion is provided at the moment arm using the magnet-coil actuator, and the shaft acts as the output. When the input is rotated clockwise, ring 3 locks to the shaft and the shaft is free to rotate clockwise relative to ring 2. Thus, the shaft rotates clockwise. When the input is rotated anti-clockwise, ring 3 is free to rotate relative to the shaft but ring 2 locks to the shaft and prohibits it from rotating anti-clockwise. Thus, the shaft remains stationary. Providing periodic clockwise plus anti-clockwise motion at the input (for example, in Fig. \ref{fig:magnet-limiter}(a), (b)) results in the shaft adding adding up all the clockwise motions and neglecting any anti-clockwise motions thus resulting in a continuous clockwise rotation motion as desired. 

The magnet guide passes through the narrow slot in the magnet limiter (see Fig. \ref{fig:magnet-limiter}(c), (d)). This serves two purposes. It ensures that the moment arm and the magnet always move in a plane perpendicular to the shaft and the base-plate. The slot also prohibits the bottom pole face of the magnet from coming out of the coil as seen in Fig. \ref{fig:magnet-limiter}(d). 

\subsubsection{Electromagnetic Actuator} 

The coil is custom made from a 25$\mu$m-thin Copper wire which is array wound $n_\text{turns} = 48\times8$ number of times. It has an inner diameter of 1.9mm, an outer diameter of 2.45mm, and a height of 1.6mm, with resistance $\approx 100\Omega$. The NdFeB magnet is of grade N52 with 1.6mm diameter and height. 

\subsection{Assembly} 

\begin{figure}[h]
\vspace{-1em}
\centering
\epsfig{file=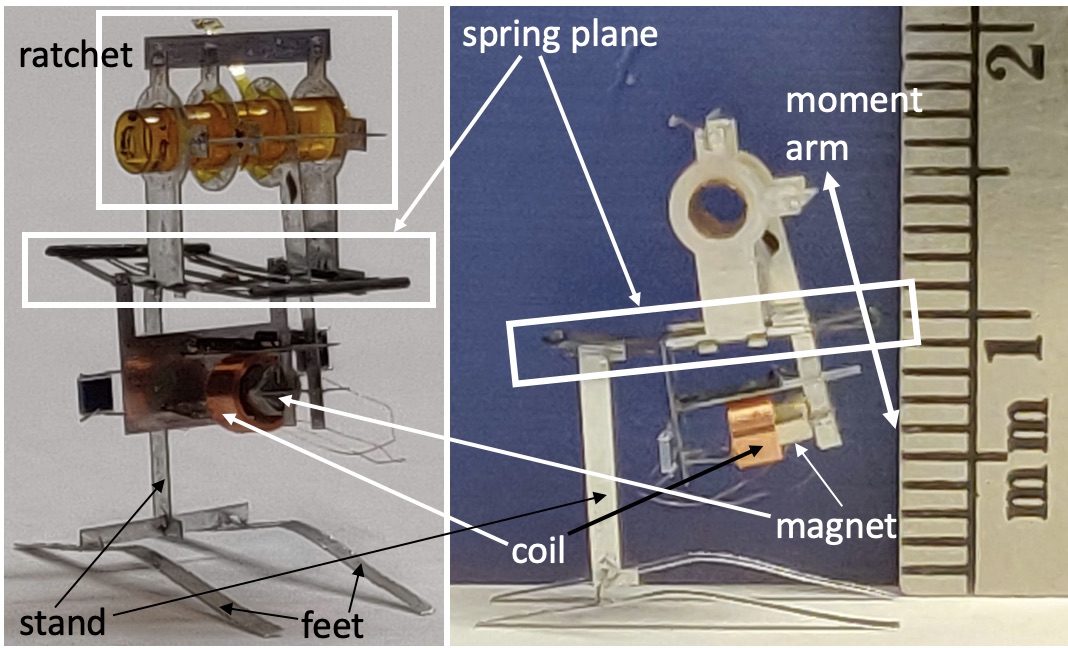,width=3.4in}
\vspace{-1.7em}
\caption{\small{Fully assembled bot compared with a millimeter ruler. }}
\vspace{-1.4em}
\label{fig:assembly}
\end{figure}

\begin{figure}[h]
\centering
\epsfig{file=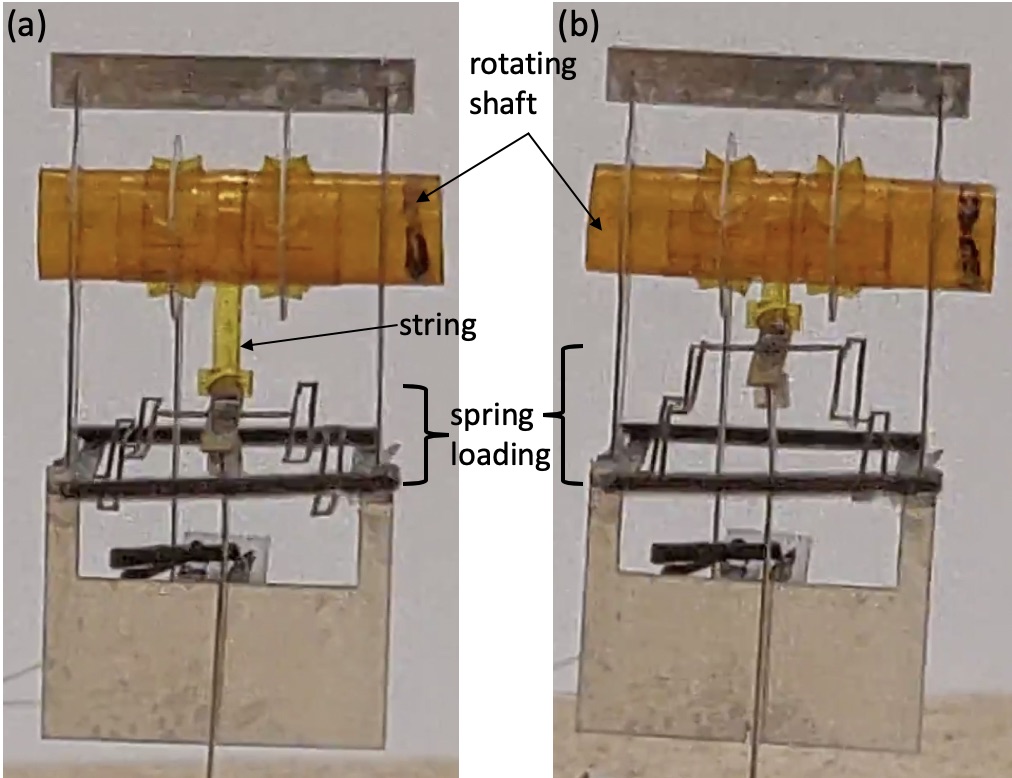,width=3.4in}
\vspace{-1.7em}
\caption{\small{Spring loading in conjunction with the double-ratchet. }}
\vspace{-1.4em}
\label{fig:spring-loading-real}
\end{figure}

The double-ratchet mechanism is glued perpendicularly to the spring designed before (see Fig. \ref{fig:assembly}). 
The fully assembled bot weights 75mg with mass distribution outlined in Table I. Note that the stand and legs are big but the rest of the bot is just a centimeter long. 
The string is attached and wound on the center of the shaft as shown in Fig. \ref{fig:spring-loading-real}. 

\begin{table}[h]
\normalsize
  \centering 
    \caption{\small Mass distribution of the $\mu$bot.}
    \vspace{-0.4em}
    \label{tab:mass}
    \begin{tabular}{|l|r|}
    \toprule 
      \textbf{Sub-component} & \textbf{Mass} \\
          \toprule
      \hline
      \multicolumn{2}{|c|}{Electrical parts}\\ 
      \hline
      Coil & 13mg \\
      Magnet + moment arm & 27mg \\
      PV cells & 2 $\times$ 1mg \\ 
        \hline
      \multicolumn{2}{|c|}{Structural parts}\\ 
      \hline
      Base-plate + supports & 9mg \\
      Ratchet tube & 9mg \\ 
      Steel spring & 2mg \\ 
      Rings + connectors & 11mg \\ 
      Stand + feet & 2mg \\ 
      \hline
      \bottomrule
      \textbf{Total} & {75mg} \\
      \bottomrule
    \end{tabular}
\vspace{-0.8em}   
\end{table}


\subsection{Starting Torque} 
Mainly two types of torques need to be overcomed to produce motion - (1) the friction torque arising from the contact between the shaft and the ring, and, (2) the torque to counteract the spring force. 

\subsubsection{To overcome friction} 
The spring pulls on the shaft which then pushes against rings 1 \& 4 (see Fig. \ref{fig:double-ratchet3}). The maximum combined contact force occurs when the spring is fully deflected and thus equals $F_\text{release}\Rightarrow$ $F_\text{contact} = $ 7.5mN. 
Assuming a friction coefficient of $\mu_s=1$, this corresponds to a starting torque of $\mu_s \cdot F_\text{contact} \cdot r_\text{shaft} = $ 7.5$\mu$Nm. A large friction coefficient is assumed since the hollow shaft is flexible and will deform under the load thus increasing its contact area with the inner perimeter of the rings. 

\subsubsection{To counteract spring force} 
Work needs to be done in order to deflect the spring and store potential energy into it. At the maximum spring deflection a string tension force of $F_\text{release}$ acts at a distance of $r_\text{shaft}$ from the center of the shaft  $\Rightarrow$ a starting torque of $F_\text{release} \cdot r_\text{shaft} = $ 7.5$\mu$Nm. 

\subsubsection{Total} 
The total estimated starting torque is thus = 15$\mu$Nm. Experimentally it is found that an applied torque of 17$\mu$Nm is sufficient to produce motion, which will now determine the minimum coil current needed to produce motion. 
Using FEA simulations we find the average magnetic field seen by the coil to be $B_\text{avg}\approx0.1$T. 
The 8mm long moment arm greatly reduces the force the coil needs to generate to produce 17$\mu$Nm of torque. $F_\text{coil} $(needed) = 17$\mu$Nm/8mm $\approx$ 2.1mN = $n_\text{turns} \cdot B_\text{avg} \cdot I_\text{coil} \cdot 2\pi r_\text{coil} \Rightarrow I_\text{coil} \approx$ 8mA. Heat loss in the coil at this current value will be $I_\text{coil}^2R_\text{coil} $ = 6.4mW, and the voltage across the coil will be $V_\text{coil}$ = 0.8V. 

\section{Experiments} 

\subsection{External power driven bot} 

\begin{figure}[h]
\vspace{-0.8em}
\centering
\epsfig{file=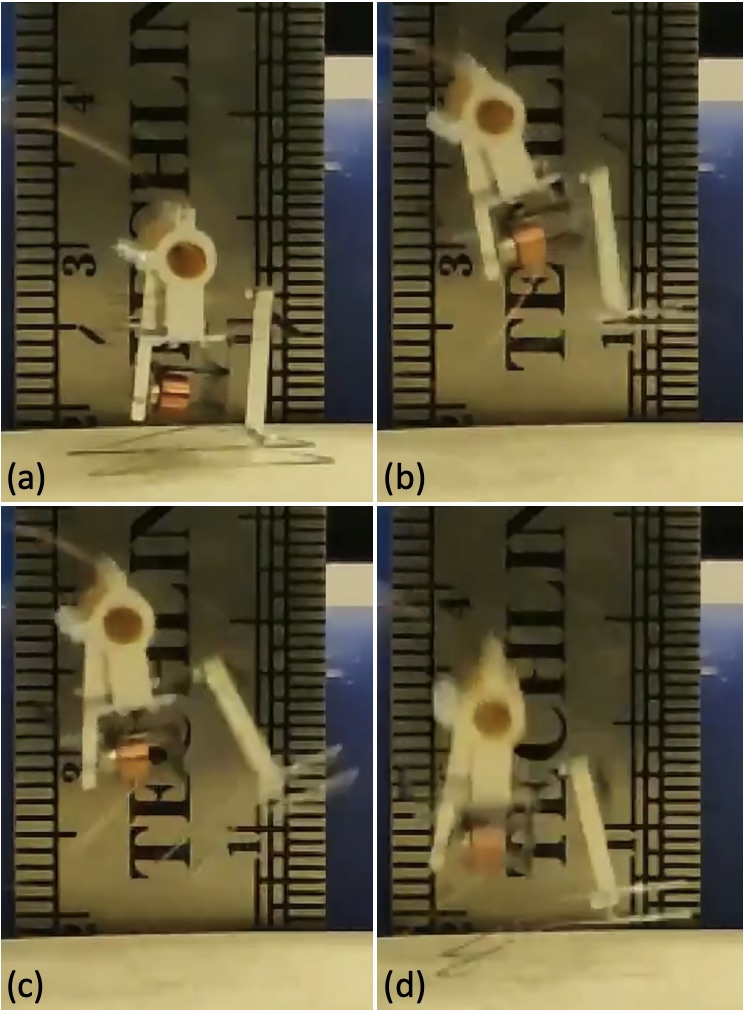,width=3.4in}
\vspace{-1.7em}
\caption{\small{Tethered jump of the bot using external power supply. (a) Before take-off. (c) Highest position. (d) Feet touchdown. }}
\vspace{-1.2em}
\label{fig:tethered-jump}
\end{figure}

The coil is powered using an external function generator and a simple square-wave voltage waveform. 
We noted that when using $V_\text{coil} = \pm 0.7$V the spring could be deflected almost completely but this voltage wasn't enough to release the magnets. A voltage of $V_\text{coil} = \pm 0.8$V was needed to passively release the magnets and cause the bot to jump. For our 75mg bot, a stored spring energy of 11.25$\mu$J should cause the bot to jump up by 15mm. In practice, however, this jump height is close to 8mm (see Fig. \ref{fig:tethered-jump}(c)) possibly due to wind resistance, inefficient spring to kinetic energy conversion \cite{churaman11}, and other device non-idealities. 

The jumping rate or the number of jumps the bot can do in a minute is determined by how fast we can load the spring. Here we operated the actuator at 20Hz which then caused the bot to jump once every 10 seconds while consuming 6.4mW of power.
This power is mostly the Joule heat loss in the coil, as the mechanical power required to overcome friction and load the spring is in tens of $\mu$Watts. Thus in theory the jumping rate could be increased by an order of magnitude while still consuming the same amount of power. 
The bot can do multiple successive jumps at this rate of 6 jumps/min without tipping over. We think the elasticity of the 2 external wires powering the coil provides this stability which prevents the bot from falling over. 

\subsection{Photovoltaics driven bot} 

Here the power source we use is a 1mm$\times$1mm infrared PV cell (MH GoPower 5S0101.4-W) that produces current when a 976nm wavelength laser light (MH GoPower LSM-010) is shone on it. The laser intensity is increased till the PV cell outputs $\approx$ 0.8V while driving a 100$\Omega$ load. 

\begin{figure}[h]
\vspace{-0.8em}
\centering
\epsfig{file=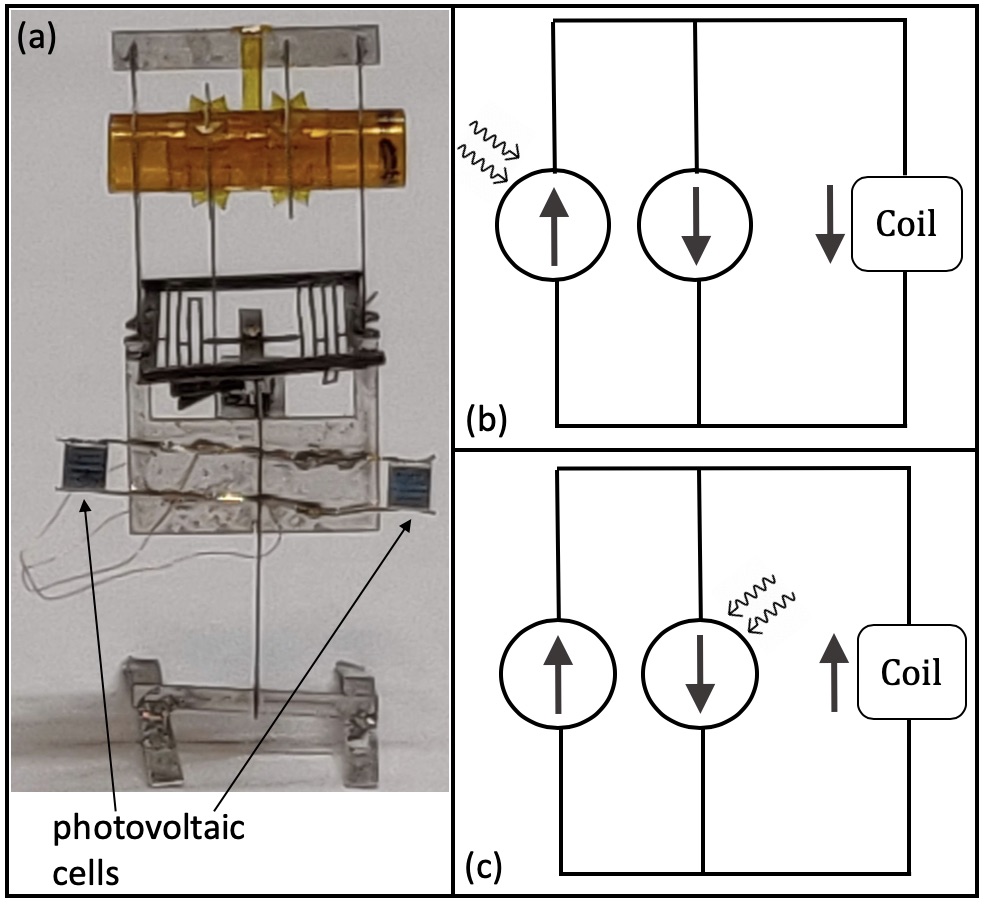,width=3.0in}
\vspace{-0.6em}
\caption{\small{Circuit with 2 PV cells for tetherless $\mu$bot operation. }}
\vspace{-1.4em}
\label{fig:circuit}
\end{figure}

\begin{figure}[h]
\centering
\epsfig{file=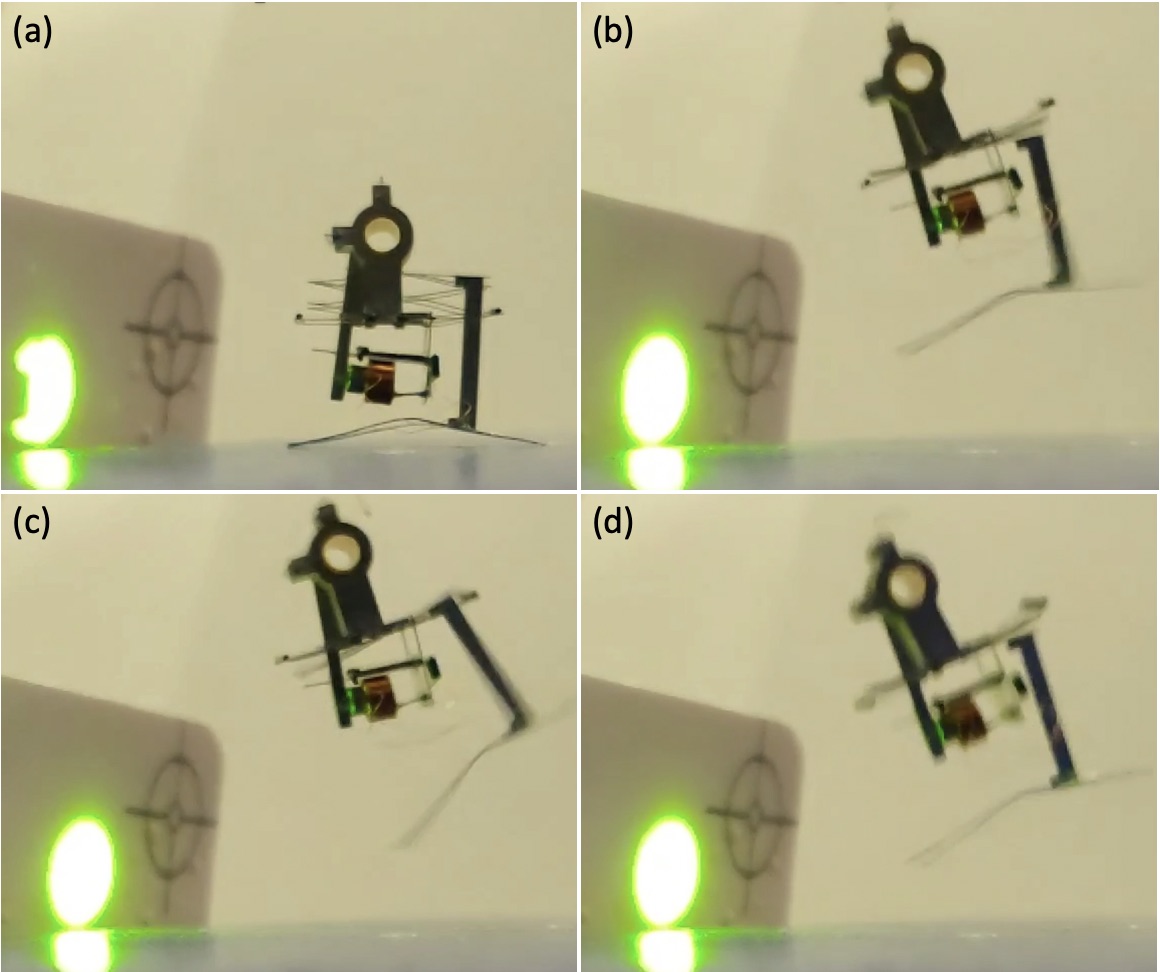,width=3.4in}
\vspace{-1.7em}
\caption{\small{Tetherless jump of the $\mu$bot using laser power. (a) Before take-off. (c) Highest position. }}
\vspace{-1.8em}
\label{fig:tetherless-jump}
\end{figure}

We connect 2 PV cells to the coil in opposite polarity as shown in Fig. \ref{fig:circuit}. Shining the laser on one PV cell makes the current go one way (see Fig. \ref{fig:circuit}(b)), while shining on the other makes it go the other way (see Fig. \ref{fig:circuit}(c)). This alternate shining of the laser is done manually to drive the actuator which then loads the spring. The laser used is an infrared laser which is invisible to the naked eye. The white card seen in Fig. \ref{fig:tetherless-jump} is an indicator card that emits green light when struck by the IR laser so that we know where the laser is pointing. 

The bot jumps the same amount as before since the extra mass of the 2 PV cells is $<$ 3mg and thus negligible compared to the bot's mass. After launching up from the ground, the bot develops an anti-clockwise angular spin (see Fig. \ref{fig:tetherless-jump}). 
This results in the bot tipping over in this direction after landing on the tip of its feet. 
This falling over could be avoided, say, by further lowering the center-of-mass of the bot and supporting the spring more symmetrically using two stands (at the front and back) instead of one to enable a more spin-less take-off. 

\section{Conclusions and Future work} 

The design of this $\mu$bot can in principle work with other small-displacement linear actuators as well. 
The moment arm of the bot can be made longer, or the shaft can be made narrower, to increase leverage and store more spring energy and thus jump higher. Further, adding a horizontal component to the launch velocity can help the bot navigate around. 
The bot can be made more `flat' (like a coin) to lower its center-of-mass and to ensure that it lands on either one of its bottom or top faces, and is thus always in the correct position to make the next jump. 

Instead of pointing the laser manually over the 2 cells, we can add an electronics unit similar to the one in \cite{rolling_ral19} to provide alternating voltage to the coil. This electronics unit can also be used with an onboard power source like a micro-cell or a supercapacitor \cite{rolling_ral19} to enable completely self-sufficient jumps. 

\section*{Acknowledgement}

The authors are grateful to get support from Commission on Higher Education (award \#IIID-2016-005)
and DOD ONR Office of Naval Research (award \#N00014-16-1-2206).


\begin{thebibliography}{99}

\bibitem{locomotion08} S. Bergbreiter, ``Effective and efficient locomotion for millimeter-sized microrobots,'' {IROS}, Nice, France, Sept. 2008. 

\bibitem{actuator_selection} M. Karpelson, G-Y. Wei, and R.J. Wood, ``A Review of Actuation and Power Electronics Options for Flapping-Wing Robotic Insects,'' {IEEE Int. Conf. on Robotics and Automation}, Pasadena, CA, May 2008. 

\bibitem{robofly18} J. James, V. Iyer, Y. Chukewad, S. Gollakota, and S.B. Fuller, ``'Liftoff of a 190 mg Laser-Powered Aerial Vehicle: The Lightest Untethered Robot to Fly,'' {IEEE Int. Conf. on Robotics and Automation}, Brisbane, Australia, May 2018. 

\bibitem{xwing19} N.T. Jafferis, E.F. Helbling, M. Karpelson, and R.J. Wood, ``Untethered Flight of an Insect-Sized Flapping-Wing Microscale Aerial Vehicle,'' {Nature} 570, 491-495, 2019. 


\bibitem{churaman11} W.A. Churaman, A. P. Gerratt and S. Bergbreiter, ``First leaps toward jumping microrobots,'' {IROS}, San Francisco, CA, USA, Sept. 2011. 

\bibitem{churaman12} W.A. Churaman, L.J. Currano, C.J. Morris, J.E. Rajkowski, S. Bergbreiter, ``The first launch of an autonomous thrust-driven microrobot using nanoporous energetic silicon,'' {J. Microelectromech. Syst.} 21: 198–205, 2012. 

\bibitem{hotplate} J. Koh, S. Jung, R.J. Wood, K. Cho, ``A Jumping Robotic Insect Based on a Torque Reversal Catapult Mechanism,'' {IROS}, Tokyo, Japan, Nov. 2013. 

\bibitem{greenspun18} J. Greenspun and K.S.J. Pister, ``First leaps of an electrostatic inchworm motor-driven jumping microrobot,'' {Hilton Head Solid-State Sensors, Actuators, and Microsystems Workshop}, Hilton Head Island, SC, June 2018. 

\bibitem{baybug18} P. Bhushan and C.J. Tomlin, ``Milligram-scale Micro Aerial Vehicle Design for Low-voltage Operation,'' {IROS}, Madrid, Spain, Oct. 2018.  

\bibitem{baybug19} P. Bhushan and C.J. Tomlin, ``Design of the first sub-milligram flapping wing aerial vehicle,'' {MEMS}, Seoul, South Korea, Jan. 2019.  

\bibitem{rolling_ral19} [Under review] P. Bhushan and C.J. Tomlin, ``An Insect-scale Self-sufficient Rolling Microrobot,'' submitted to Robotics and Automation Letters (RA-L) 2019. 

\end{thebibliography}
\end{document}